\documentclass[runningheads]{llncs}

\usepackage{eccv}

\usepackage{eccvabbrv}
\usepackage{capt-of}  % 提供 \captionof 命令
\usepackage{graphicx}
\usepackage{booktabs}
\usepackage{pgfplots}
\pgfplotsset{compat=1.18}
\usepackage[table]{xcolor}  % 确保导言区有这行

\usepackage{pifont}
\usepackage{multirow}
\usepackage[accsupp]{axessibility}  % Improves PDF readability for those with disabilities.

\usepackage{hyperref}

\usepackage{orcidlink}

\begin{document}

% ---------------------------------------------------------------
% TODO REVIEW: Replace with your title
\title{UBone3D: Physics-Rectified Conditional Flow Matching for Anatomical 3D Shape Completion from Ultrasound}

% TODO REVIEW: If the paper title is too long for the running head, you can set
% an abbreviated paper title here. If not, comment out.
\titlerunning{UBone3D}

% TODO FINAL: Replace with your author list. 
% Include the authors' OCRID for the camera-ready version, if at all possible.
\author{Weiying Chen\inst{1}\orcidlink{0009-0007-9212-0745} \and
Yuchong Gao\inst{2}\orcidlink{0009-0001-2585-9319} \and
Siyuan Li\inst{1}\orcidlink{0009-0001-3990-7502} \and Marek Reformat\inst{1}\orcidlink{0000-0003-4783-0717} \and Rui Zheng\inst{2}\orcidlink{0000-0003-0391-2454} \and Edmond Lou\inst{1}\orcidlink{0000-0002-7531-8377}}

% TODO FINAL: Replace with an abbreviated list of authors.
\authorrunning{W.~Chen et al.}
% First names are abbreviated in the running head.
% If there are more than two authors, 'et al.' is used.

% TODO FINAL: Replace with your institution list.
\institute{University of Alberta, Edmonton, AB, Canada \\\email{\{weiying3,elou\}@ualberta.ca}\and
ShanghaiTech University, Shanghai, China}

\maketitle

\begin{abstract}
Three-dimensional ultrasound (US) is a safe, radiation-free complementary modality to CT and X-rays for longitudinal monitoring, yet its segmentation-derived partial point clouds are extremely artifact-laden. Consequently, it is challenging to recover a clean and complete anatomical structure from such US point clouds. In this paper, we present UBone3D, a novel framework based on physics-rectified conditional flow matching (CFM) that performs point cloud completion directly from partial US observations. UBone3D models deterministic physics artifacts (e.g., surface thickening, streaking, dropouts) via a simulated physics proxy, and introduces test-time physics rectification to steer the shape completion. At inference, the completion is jointly steered by two decoupled forces: (1) anatomical plausibility enforced by a CT-trained generative shape prior, \textbf{BoneFM}, and (2) physics consistency enforced by \textbf{USimNet} in the ultrasound formation space. Extensive experiments on simulated and in-vivo data demonstrate significant improvements in reconstruction accuracy and anatomical fidelity over existing baselines. Project page: \url{https://answerrtx.github.io/UBone3D-Proj/}.

\end{abstract}

\section{Introduction}
\label{sec:intro}

Reconstructing accurate three-dimensional (3D) bone geometry directly from ultrasound scans remains a fundamental unsolved challenge in medical imaging. A robust solution would enable the next generation of radiation-free, real-time 3D applications, from intraoperative navigation to affordable long-term monitoring of skeletal deformities. This is especially important for adolescent idiopathic scoliosis (AIS), where patients often need repeated CT or X-rays over years, resulting in high cumulative radiation exposure. In current clinical practice, 3D spine models are typically generated by aggregating tracked 2D B-mode ultrasound frames according to their spatial poses and volumetrically interpolating the data using Voxel Nearest Neighbor (VNN)~\cite{chen2021improvement} for downstream assessment. VNN voxel stacking is a lossy discretization that blurs thin cortical surfaces and propagates physics-induced dropouts and reverberation into large empty or spurious regions, further obscuring the true bone geometry. 

\begin{figure}[t]
    \centering
    \includegraphics[width=1\linewidth]{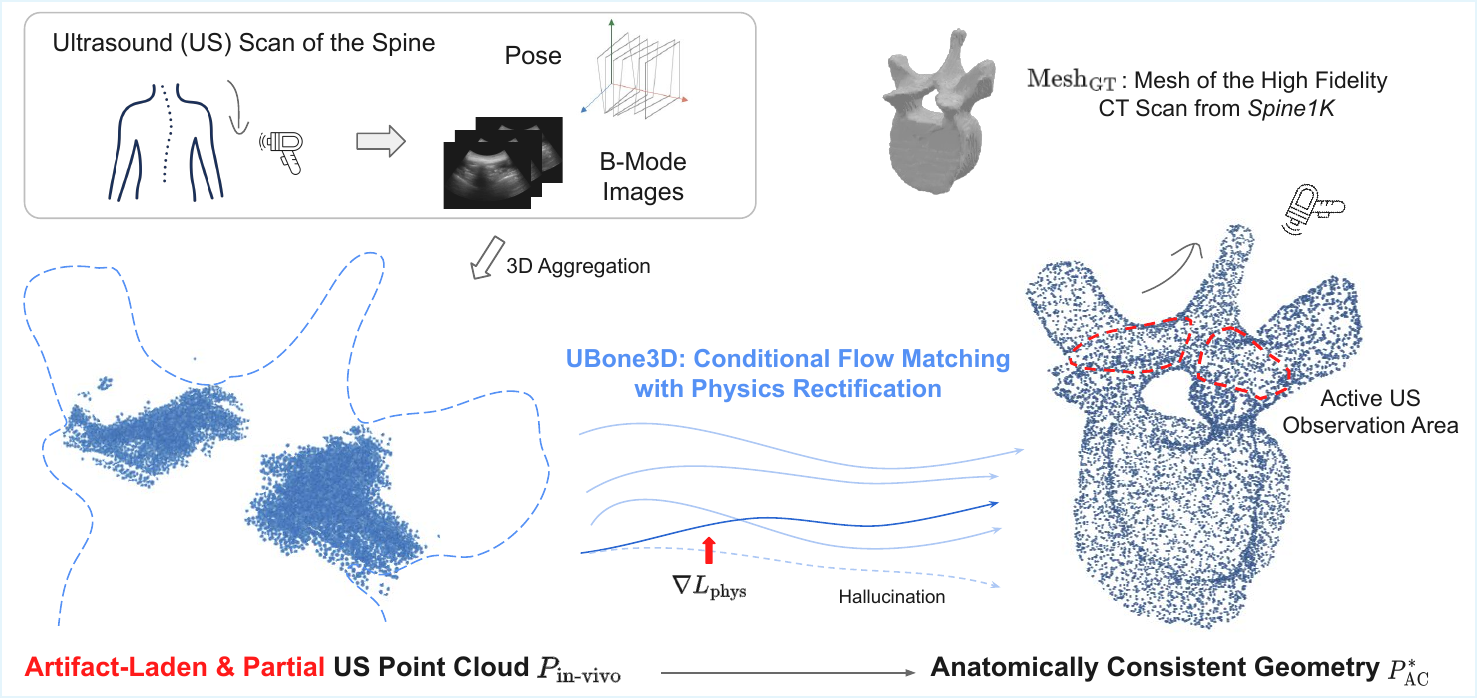}
    \caption{\textbf{Physics-rectified shape completion bridges the gap between ultrasound (US) observations and complete anatomy.} (Left) Segmentation-derived in-vivo US point clouds suffer from severe acoustic artifacts and only partially cover the underlying bone anatomy. (Middle) UBone3D leverages conditional flow matching (CFM) with a differentiable physics proxy to steer the generative trajectory, constraining completion within the physical bounds of ultrasound formation. (Right) The reconstructed anatomical completion ($\text{P}_{\text{AC}}$) achieves global structural integrity while maintaining high local fidelity to the observed ultrasound signals (red regions).}
    \label{teaser}
\end{figure}

Ideally, the complete skeletal structure could be recovered from cleanly segmented partial structures; however, achieving perfect segmentation is often impractical in real clinical settings. In this paper, we choose the point-cloud representation to directly model and recover the underlying geometry from \textbf{imperfect segmentation-derived ultrasound input}. Recovering an anatomically clean and complete 3D structure from such input is inherently challenging: beyond the missing topology common to any partial observation, the artifact-laden ultrasound point cloud extracted from upstream segmentation exhibits a large gap relative to the desired anatomically complete geometry, as shown in Fig.~\ref{teaser}. Importantly, this gap is \emph{not merely one of random noise, but one of deterministic physics}: finite beamwidth, reverberation, and signal dropout caused by the probe's limited aperture (FOV) together distort the observed signals, leading to incomplete and highly nonuniform point clouds.

Recent 3D reconstruction techniques for ultrasound imaging~\cite{gafencuShapeCompletionDark2024,wysockiUltraNeRFNeuralRadiance2023,chenNeuralImplicitSurface2024b,ultron2025} focused on recovering a clean bone surface of the superficial bone boundaries, assuming access to clean geometric or intensity data, neglecting that in-vivo ultrasound scans do not meet this standard. On the other hand, general 3D completion models were trained on idealized, clean geometric data~\cite{yu2021pointr,zhu2023svdformer,vahdat2022lion,wei2025pcdreamer}. The foundational knowledge, or priors, used by these generative methods often comes from datasets like ShapeNet~\cite{chang2015shapenet} or natural images, giving them no understanding of the acoustic physics inherent to ultrasound. 

In this paper, we introduce \textbf{UBone3D}, a novel physics-rectified conditional flow matching framework that casts the recovery of complete anatomical geometry from partial ultrasound as an observation-guided denoising and shape completion problem. The key insight is to decouple two complementary objectives, anatomical plausibility and physical consistency, which are respectively handled by BoneFM and USimNet and synergized during test-time inference. Our main contributions include:

\begin{itemize} 
\item UBone3D enables the recovery of anatomically consistent complete 3D structures from artifact-laden, partial ultrasound observations by synergizing the generative anatomy prior (BoneFM) with test-time physical rectification, without relying on clean or perfectly segmented inputs.

\item We introduce USimNet, a differentiable physics proxy that models acoustic artifacts (e.g., bone shadowing and streaks), supporting an explicit, gradient-based rectification mechanism that aligns the evolving generative trajectory with physical observation constraints during inference.

\item Our approach demonstrates strong generalization to unseen viewing angles on simulated data, and remains robust on zero-shot in-vivo data, where pure geometric baselines fail.
\end{itemize}

\section{Literature Review}
\label{sec:literature}
 \subsection{3D Point Cloud Completion and Generation}

Early 3D point cloud completion methods largely followed encoder–decoder designs. Representative works such as PCN~\cite{yuan2018pcn} and GRNet~\cite{xie2020grnet} employed coarse-to-fine generation to progressively recover missing geometry. Transformer-based models later advanced this field by encoding partial inputs into point proxy sequences and leveraging geometry-aware attention to capture long-range dependencies for complete shape decoding~\cite{yu2021pointr}. Subsequent methods further refined this paradigm through hierarchical point-wise deconvolution~\cite{xiang2021snowflakenet}, view-structure disentanglement~\cite{zhu2023svdformer}, or specialized guidance mechanisms like seed expansion~\cite{zhou2022seedformer} and proxy-based alignment~\cite{li2023proxyformer}. On the other hand, generative approaches (e.g., ContextualCompletion~\cite{chu2025digging}, PCDreamer~\cite{wei2025pcdreamer}) formulated completion as conditional sampling from prior distributions. Recent works also extended completion to cross-modal settings~\cite{mao2025dmf, du2024cdpnet, luo2025rethinking,du2025superpc,Wang_2025_CVPR}, which took auxiliary inputs to provide explicit guidance and alleviate the ill-posed nature of shape recovery. Notably, SuperPC~\cite{du2025superpc} unified completion from partial observations with other restoration tasks, including upsampling, denoising, and colorization. 

In clinical ultrasound scenarios, however, acquiring perfectly aligned data is highly impractical due to tissue deformation, probe pressure, and respiratory motion during scanning. Driven by similar practical constraints, some works have explored unpaired point cloud completion, e.g., USSPA~\cite{USSPA} and UOT-UPC~\cite{uot-upc}. However, in ultrasound-to-anatomical comprehensive bone completion, the key challenge lies in the deterministic, physics-driven ultrasound artifacts that systematically distort the observations.
 
\subsection{Physics-Guided Generation and Neural Surrogates}

 Recent physics-guided frameworks have used learnable models with real-world physics simulations to enhance the generation quality. For example, PIDM~\cite{bastekphysics} imposed governing laws as the training-time regularizer to ensure that predictions remain physically self-consistent. In contrast, PhysDiff~\cite{yuan2023physdiff} incorporated a human motion simulator at inference time as guidance, progressively refining generated samples to enforce physical consistency. Such physics-guided strategies have been extended to different domains, including geometrical design~\cite{Jiang_2025_ICCV, Zheng_2025_ICCV, giannone2023aligning}, and animation~\cite{Xie_2025_CVPR}. In our practice, the physics prior, USimNet, serves as a lightweight learned surrogate that captures physics effects while maintaining the gradient flow necessary for test-time rectification.

\subsection{Medical Shape Reconstruction for Bone} 
Traditionally, Statistical Shape Models (SSMs)~\cite{rusli2020statistical,aubertAutomated3DSpine2019a,wai2023statistical,cootes2024statistical} were used to perform 3D reconstruction by learning a parametric shape representation. More recently, the 3D medical shape reconstruction has shifted toward using Implicit Neural Representations (INRs), e.g., Signed Distance Functions (SDFs) and occupancy fields for the surface reconstruction~\cite{chenNeuralImplicitSurface2024b, chenRoCoSDFRowColumnScanned2024a, ultron2025, AMIRANASHVILI2024103099}. Although these works excelled at capturing complex topologies, they typically rely on clean and relatively complete observations. In the context of ultrasound bone imaging, volumetric representations were often reconstructed via Voxel Nearest Neighbor (VNN)~\cite{chen2021improvement} interpolation according to spatial poses in clinical practice. Alternatively, point-cloud-based approaches have been proposed to register ultrasound with CT for intraoperative navigation in spine surgery~\cite{Li2025}. While these methods do not provide global anatomy, Gafencu \etal~\cite{gafencuShapeCompletionDark2024,gafencu2025us} explored leveraging ultrasound-physics-aware ray-casting to generate synthetic partial vertebrae point clouds for vertebral shape completion. Although accounting for basic ultrasound artifacts, their method was based on a variational autoencoder (VAE)~\cite{kingma2013auto}, which inherently maps uncurated, complex real-world artifacts directly into the latent space. This may lead to shape collapse on in-vivo data, whose distribution differs from that of the simulated data. In contrast, our work formulates US 3D completion as a conditional generative task. By leveraging a large-scale CT dataset (Spine1K~\cite{deng2021ctspine1k}) to form the anatomy prior, UBone3D recovers globally consistent geometry while accounting for physics-induced artifacts.

% \input{content/3method}
% ===============================
\section{Methodology}

% ===============================
\subsection{Preliminary: Optimal Transport Flow Matching}
\label{sec:preliminary_otfm}

 In this work, we build our generative prior~\textbf{BoneFM} based on Optimal Transport Flow Matching (OT-FM)~\cite{lipman2022flow, liu2022flow}, which provides a gradient path to transport a simple base distribution to a complex data distribution. 

Let $x \in \mathbb{R}^{3N}$ denote the vectorized representation of a point cloud 
$P = \{p_i\}_{i=1}^N \subset \mathbb{R}^3$, and $t \in [0, 1]$ be the continuous time variable. We define the distributions as:
\begin{equation}
x_0 \sim p_0(x) = \mathcal{N}(0, \mathbf{I}), \qquad x_1 \sim p_1(x), \quad \text{where } p_1(x) \approx p_{\text{AC}}(x),
\end{equation}
where $p_0(x)$ is a standard Gaussian noise distribution and $p_{\text{AC}}(x)$ represents the target distribution of clean and anatomically consistent bone geometries. 

To obtain a highly efficient transport path, OT-FM constructs a path under the OT coupling that connects $x_0$ and $x_1$ via linear interpolation~\cite{albergo2023building}:
\begin{equation}
x(t) = t x_1 + (1 - t) x_0.
\label{eq:ot_path}
\end{equation}
Taking the time derivative of Eq.~\ref{eq:ot_path} yields the optimal transport velocity:
\begin{equation}
u(x(t), t)=x_1-x_0
\label{eq:ot_velocity}
\end{equation}
Such straight-trajectory formulations are highly desirable as they accelerate ordinary differential equation (ODE) integration during inference~\cite{kornilov2024optimal}. A neural velocity field $v_\theta(x, t)$ can be trained to approximate $u(t)$ using a simple regression objective:
\begin{equation}
\mathcal{L}_{\text{FM}} = \mathbb{E}_{t \sim \mathcal{U}(0,1),\, x_0 \sim p_0,\, x_1 \sim p_1} \left[ \left\| v_{\theta}(x(t), t) - (x_1 - x_0) \right\|_2^2 \right].
\label{eq:otfm_loss}
\end{equation}
In our UBone3D, this formulation is later extended to a conditional velocity field, 
where conditioning variables, i.e., the ultrasound observation, reshape the transport trajectory toward the observation-consistent posterior manifold. Throughout this work, 
we use $x$ to denote the vectorized representation of the target anatomy point cloud 
$P_{\text{AC}}$, and $y$ to denote the ultrasound observation point cloud $P_{\text{in-vivo}}$ or $P_{\text{phys-full}}$.
% ===============================
\subsection{Framework Overview}
\label{sec:framework_overview}

\begin{figure*}[t]
    \centering
    \includegraphics[width=1\linewidth]{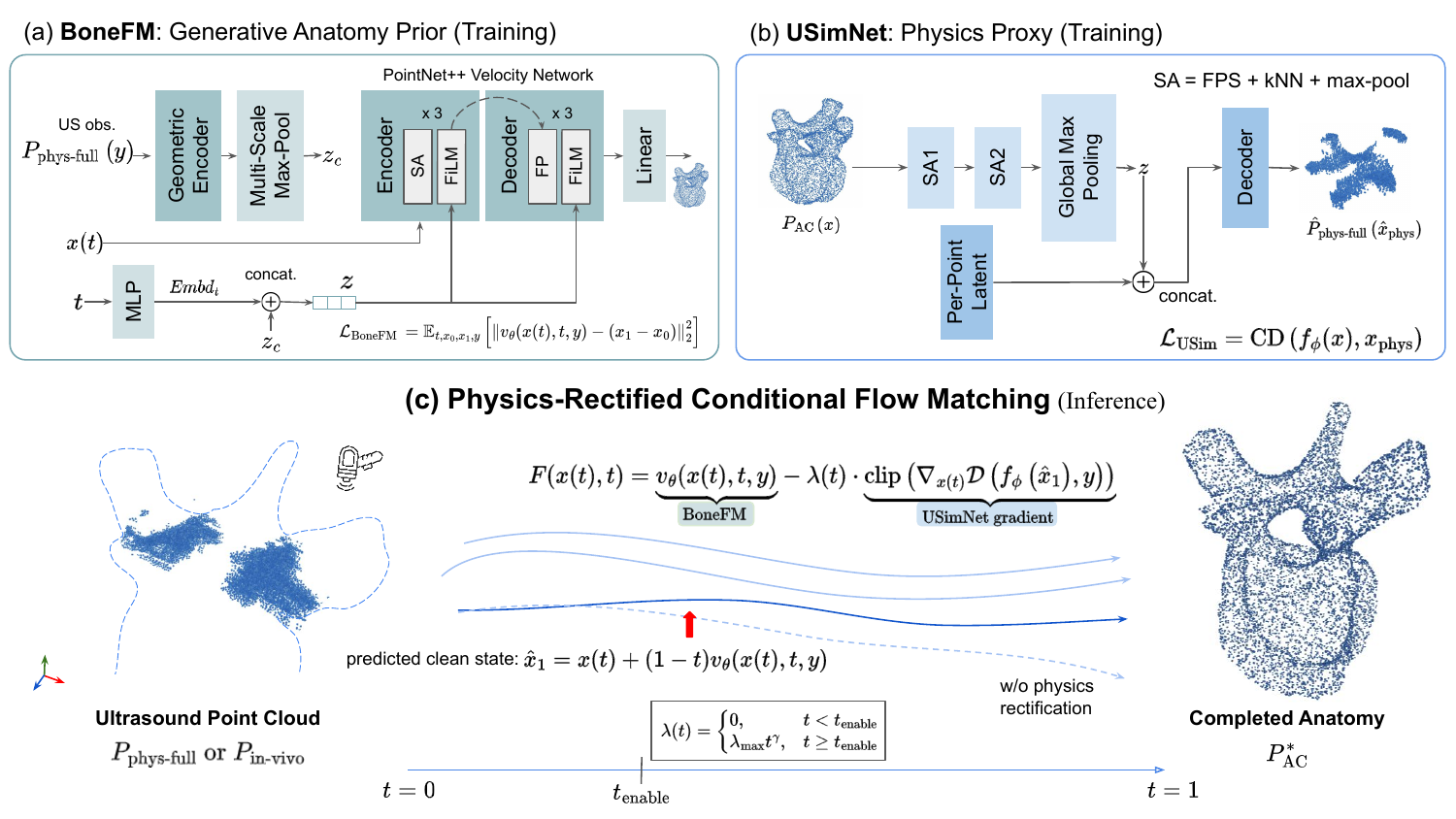}
\caption{%
    \textbf{Overview of the UBone3D framework.}
    \textbf{(a) BoneFM Training:} The generative anatomy prior is learned by training 
    a velocity field $v_\theta$ to transport a base distribution toward the manifold 
    of clean CT anatomy $P_{\text{AC}}$.
    \textbf{(b) USimNet Training:} Our differentiable physics proxy $f_\phi$ learns 
    to project a complete anatomical point cloud $P_{\text{AC}}$ into its 
    ultrasound-style counterpart $P_{\text{phys-full}}$.
    \textbf{(c) Rectified Inference:} Given the artifact-laden, partial observation, we obtain an anatomically consistent reconstruction 
    $P_{\text{AC}}^*$ with the rectified inference. $v_\theta$ here denotes the CFG velocity.
    %The trajectory is jointly determined by the BoneFM backbone and a physics-rectification gradient $\nabla_{x(t)} \mathcal{D}$ computed via USimNet on the predicted clean state~$\hat{x}_1$.%
}
\label{pipe}
\end{figure*}

UBone3D aims to infer an \textbf{anatomically consistent completion} $x^*$ that recovers the full vertebral geometry from the artifact-laden and partial \textbf{in-vivo} ultrasound observation $y$. We formulate this observation-guided shape completion as a probabilistic decomposition:
\begin{equation}
x^* \sim p(x \mid y) \propto p(y \mid x) p(x).
\label{eq:bayes_decomp}
\end{equation}
The final completed point cloud is obtained by interpreting the state $x^*$ as a 3D point cloud $P_{\text{AC}}^* = \{p_i^*\}_{i=1}^N$.
As illustrated in Fig.~\ref{pipe}, UBone3D integrates two modules during inference: a generative backbone and a physics rectifier:

\begin{itemize}
    \item \textbf{Generative Anatomy Prior (BoneFM):} BoneFM parameterizes a conditional velocity field $v_\theta(x(t), t, y)$ optimized via the OT-FM objective (Eq.~\ref{eq:otfm_loss}). Given an observation $y$, it transports samples from a standard Gaussian base distribution toward the clean anatomical manifold $p_{\text{AC}}$.
    \item \textbf{Physics Proxy (USimNet):} We model $p(y \,|\, x)$ with USimNet $f_\phi$, a differentiable surrogate network, to \textit{replicate the geometric distribution of in-vivo observations by learning from physics-based simulation}. Given a candidate complete shape $x$, it acts as a data-driven forward model that maps the clean anatomy into the ultrasound domain: $\hat{x}_\text{phys} = f_\phi(x)$. 
\end{itemize}

At test time, BoneFM and USimNet jointly drive a \textbf{physics-rectified} conditional flow matching (Sec.~\ref{sec:inference_guided_flow}), where the generative anatomy prior provides a shape basis, and USimNet supplies explicit observation-consistency gradients to steer the flow toward the posterior manifold.

% ===============================
\subsection{Generative Anatomy Shape Prior: Conditional BoneFM}
\label{sec:otfm_prior}

 We train our generative anatomy prior, \textbf{BoneFM}, on an anatomically clean dataset, and model it as a \textit{conditional} velocity field, as established in Sec.~\ref{sec:preliminary_otfm}. 

\subsubsection{Conditional Velocity Parameterization.}
BoneFM employs a PointNet++-based~\cite{qi2017pointnet++} architecture enhanced with Feature-wise Linear Modulation (FiLM)~\cite{perez2018film}. The partial ultrasound observation $y$ is first processed by a geometric encoder to extract a global context vector. Meanwhile, the continuous integration time $t$ is mapped to a high-dimensional sinusoidal embedding. These conditional signals are then fused and injected into the intermediate layers of the velocity network via FiLM, which applies affine transformations (scale and shift) to the intermediate point features. This allows the network to dynamically adjust its spatial receptive field based on the ultrasound observation.

\subsubsection{Training Objective.}
Following the optimal transport path defined in Eq.~\ref{eq:ot_path}, BoneFM minimizes a conditional adaptation of the regression loss established in Eq.~\ref{eq:otfm_loss}. The network is optimized to predict the constant target velocity $u(t) = x_1 - x_0$ conditioned on the partial input $y$:
\begin{equation}
\mathcal{L}_{\text{BoneFM}} = \mathbb{E}_{t, x_0, x_1, y} \left[ \left\| v_{\theta}\big(x(t), t, y\big) - (x_1 - x_0) \right\|_2^2 \right],
\label{eq:bonefm_loss}
\end{equation}
where the target state $x_1$ corresponds to $P_{\text{AC}} = \{p_i\}_{i=1}^N \subset \mathbb{R}^3$, the anatomically consistent ground-truth point cloud, and $y \in \mathbb{R}^{3 M}$ denotes the simulated partial ultrasound observation point cloud. By mapping the standard Gaussian base distribution to clean anatomical shapes conditioned on ultrasound observation $y$, BoneFM establishes a robust and smooth topological foundation for the subsequent test-time shape guidance. To enable classifier-free guidance (CFG) at inference, 
the conditioning signal $y$ is randomly replaced by a null token 
during training with probability $p_\text{drop}$.
% ===============================
\subsection{Differentiable Physics Proxy: USimNet}
\label{sec:usimnet}

\subsubsection{Physics Simulation.}
\label{sec:physics_sim}

To obtain physically grounded supervision for ultrasound appearance, we employ a customized ray-based simulator based on the open-source \textit{PyMUST}~\cite{bernardino2024pymust}, operating directly on anatomically clean meshes.
For each mesh, we construct an orthonormal probe basis $(u,v,w)$ and cast a dense set of rays from a start plane along the probe direction $-w$. Specifically, we first enforce a sector-shaped \textbf{field-of-view (FOV)} by clipping rays whose angular deviation from the probe apex exceeds a predefined aperture, e.g., $73^\circ$. The simulated probe is tilted around the world $X$-axis, e.g., by $+10^\circ$, to simulate realistic probe handling and incidence variations. We also model key ultrasound artifacts in terms of the following features through analytic filters and augmentations:

\begin{itemize}
    \item \textbf{Angle Gating and Shadowing (Shadow):} Suppressing signal returns behind strong cortical reflections within a set lateral radius and depth margin;
    \item \textbf{Depth Windowing and Attenuation (Atten):} Constraining the effective imaging depth and decaying intensities;
    \item \textbf{Thick-Surface Augmentation (Thick):} Adding samples slightly inside the bone surface along the inward normal to mimic thick reflective surfaces;
    \item \textbf{Axial Streak Artifacts (Streak):} Duplicating hits along the ray direction to emulate streak-like artifacts;
    \item \textbf{Anti-Grid Sampling (Anti-Grid):} Adding stochastic variability (Poisson sampling, jitter, and noise) to reduce grid regularity.
\end{itemize}
\noindent After applying these physically motivated filters, we obtain a simulated ultrasound-style point cloud $P_{\text{phys-full}}$ (as detailed in Sec.~\ref{sec:dataset_sim}). This simulated point cloud is used as supervision for training our differentiable proxy,~\textbf{USimNet}. 

\subsubsection{USimNet Architecture and Training.}
\label{sec:usimnet_train}
Although the ray-based simulator described in Sec.~\ref{sec:physics_sim} 
produces physically grounded supervision, it is non-differentiable 
and computationally prohibitive for test-time gradient computation. 
Therefore, we introduce \textbf{USimNet}, a lightweight, differentiable surrogate network $f_{\phi}$, to learn a data-driven forward mapping that approximates complex physical artifacts (e.g., acoustic shadows, limited field-of-view) with the simulator's output. It employs a PointNet++-based~\cite{qi2017pointnet++} encoder-decoder architecture. The encoder captures hierarchical geometric patterns with Set Abstraction (SA) layers. These features are then symmetrically pooled into a global context vector and processed by an MLP-based decoder. 

Given the complete anatomically consistent point cloud $P_{\text{AC}} = \{p_i\}_{i=1}^N \subset \mathbb{R}^3$, represented as 
$x \in \mathbb{R}^{3N}$, USimNet learns a non-linear spatial transformation to project the input geometry into the simulated ultrasound point cloud $\hat{P}_{\text{phys-full}} \subset \mathbb{R}^3$:
\begin{equation}
\hat{x}_{\text{phys}} = f_{\phi}(x),
\end{equation}
where $\hat{x}_{\text{phys}} \in \mathbb{R}^{3N'}$ denotes the simulated ultrasound point cloud $\hat{P}_{\text{phys-full}}$ for geometric evaluation.

\subsubsection{Training Objective.}
USimNet is trained by minimizing the symmetric Chamfer Distance (CD), which can be formulated as:
\begin{equation}
\mathcal{L}_{\mathrm{USim}}=\mathrm{CD}\left(f_\phi(x), x_{\mathrm{phys}}\right)
\end{equation}
where $x$ and $x_{\text{phys}}$ 
denote vectorized representations of 
$P_{\text{AC}}$ and 
$P_{\text{phys-full, train}}$, respectively.

In summary, USimNet provides a differentiable mapping $P_{\text{AC}} \rightarrow \hat{P}_{\text{phys-full}}$. By acting as a structural \emph{physics proxy}, it can supply explicit observation-consistency gradients during the test-time flow inference.
% ===============================
\subsection{Inference: Physics-Rectified Conditional Flow Matching}
\label{sec:inference_guided_flow}

Given the artifact-laden and partial ultrasound observation $y$, we reconstruct the complete anatomy $P_{\text{AC}}^*$ via physics-rectified conditional flow matching. 

Starting from pure noise $x(0)\sim\mathcal{N}(0,\mathbf{I})$, we numerically solve the flow from $t=0$ to $t=1$ with Heun's second-order method for a stable integration of the generative prior and the physics correction. At each step $\Delta t$, the state is updated as:
\begin{equation}
\begin{aligned}
k_1 &= F\big(x(t), t\big), \\
k_2 &= F\big(x(t) + \Delta t\, k_1,\; t+\Delta t\big), \\
x(t+\Delta t) &= x(t) + \frac{\Delta t}{2}\big(k_1 + k_2\big),
\end{aligned}
\label{eq:guided_heun}
\end{equation}
where 
\begin{equation}
F(x(t), t)
=
% v_\theta\!\big(x(t), t, y\big)
v_{\mathrm{cfg}}\!\big(x(t), t, y\big)
-
\lambda(t)\,\operatorname{clip}\!\left(
\nabla_{x(t)} 
\mathcal{D}\!\big(
f_\phi(\hat{x}_1(x(t), t)),\, y
\big)
\right),
\label{eq:guided_vectorfield}
\end{equation}
where $v_{\mathrm{cfg}}(x,t,y) = (1+w)\,v_\theta(x,t,y) - w\,v_\theta(x,t,\varnothing)$ 
is the classifier-free guidance velocity with weight $w$. $\operatorname{clip}(\cdot)$ denotes a per-point gradient normalization that prevents numerical instability caused by outlier gradients. All gradients are computed via automatic differentiation. At intermediate state $x(t)$, we predict the clean target shape before evaluating the physics proxy $f_\phi$ since $x(t)$ does not lie on the clean data 
manifold:
\begin{equation}
% \hat{x}_1\big(x(t), t\big) = x(t) + (1-t)\,v_\theta\big(x(t), t, y\big),
\hat{x}_1\big(x(t), t\big) = x(t) + (1-t)\,v_{\mathrm{cfg}}\!\big(x(t), t, y\big).
\label{eq:pred_x1}
\end{equation}

To form an explicit rectification gradient on BoneFM, we apply a geometric discrepancy $\mathcal{D}$ with the directed Chamfer Distance:
\begin{equation}
\mathcal{D}\!\left(
f_\phi\big(\hat{x}_1(x(t), t)\big),\, y
\right)
=
\operatorname{CD}_{\text{directed}}\!\left(
y,\,
f_\phi\big(\hat{x}_1(x(t), t)\big)
\right).
\end{equation}
The directed CD is used to avoid penalizing regions that are not covered by $y$.

\textbf{Observation Anchoring.}
During integration, we hold a subset of $N_\text{vis}$ observation points on their exact OT-FM path $x_\text{vis}(t)=t\,y_\text{vis}+(1-t)\,z_\text{vis}$ (so $x_\text{vis}(1)=y_\text{vis}$), letting the rectification complete only the unobserved region.

\textbf{Time-dependent Guidance Schedule.}
Applying physical gradients in early, high-noise stages might corrupt the global topology. Therefore, we restrict USimNet's intervention to the late generative phase using a truncated polynomial schedule:
\begin{equation}
\lambda(t) = 
\begin{cases} 
0, & \text{if } t < t_{\text{enable}}, \\
\lambda_{\text{max}} t^\gamma, & \text{if } t \ge t_{\text{enable}},
\end{cases}
\end{equation}
where $t_{\text{enable}}$ represents the activation threshold and $\gamma$ represents the scaling factor. Detailed parameters are provided in Sec.~\ref{sec:implement}.

\section{Experimental Setup} 

\subsection{Dataset} 
\textbf{Simulation Data.}
\label{sec:dataset_sim}
We curated \textbf{Spine1K-PC}, a large-scale simulated point cloud dataset derived from Spine1K~\cite{deng2021ctspine1k}. To establish consistent geometric ground truths (GTs), we extracted individual vertebrae from the raw volumetric scans and applied the Marching Cubes algorithm to generate the ground-truth meshes ($\mathcal{M}_{\text{GT}}$). Subsequently, we uniformly sampled $N=8,192$ points from $\mathcal{M}_{\text{GT}}$ to obtain the \textbf{Anatomically Consistent Point Cloud} ($P_{\text{AC}}$). This serves as the complete anatomical reference that our model aims to recover. This dataset was split into 5,502 samples for training and 1,373 samples for testing. We generated four distinct variants of partial ultrasound observations, all based on the same field-of-view cut (FOV-Cut) and vertebral body removal (Body-Cut):

\begin{description}
    \item[1) Pure Geometric Train ($P_{\text{geo, train}}$):] We simulated a restricted observation by taking the top-35\% points (Top 35\%) at a predefined tilted angle of the probe on $P_{\text{AC}}$. This serves as the most basic, artifact-free partial input.
    
    \item[2) Simple Physics Train ($P_{\text{phys-simp, train}}$):] We applied PyMUST~\cite{bernardino2024pymust} on $\mathcal{M}_{\text{GT}}$ to simulate minimal physics effects, i.e., first-return ray casting (Ray) and spinous-process removal (SP-Cut) under the same probe angle as $P_{\text{geo}}$.
    
    \item[3) Full Physics Train ($P_{\text{phys-full, train}}$):] We used PyMUST to simulate comprehensive acoustic physics, including acoustic shadowing (Shadow), depth windowing and exponential attenuation (Atten), thick-surface responses (Thick), axial streak artifacts (Streak), and anti-grid sampling perturbations (Anti-Grid: Poisson + jitter + subsampling + noise), under the same probe angle as $P_{\text{geo}}$ on the training set.
    
    \item[4) Full Physics Test ($P_{\text{phys-full, test}}$):] To evaluate generalization capability, we simulated full physics effects on 1,373 test samples using a \textit{novel, strictly unseen} probe angle.
\end{description}

\textbf{In-Vivo Data.} With ethics approval and informed consent, we retrospectively collected 5 cases from healthy adult volunteers using a Clarius transducer (Clarius Mobile Health, BC, Canada) following a standard protocol. We then reconstructed a 3D vertebral point cloud of the full spine for each case after segmentation on the B-mode images. Finally, we extracted 24 vertebrae in total from the thoracic (20) and lumbar (4) regions exclusively for the sim-to-real test. The resulting point clouds \textbf{$P_{\text{in-vivo, test}}$} were normalized into a canonical space. 

\subsection{Implementation Details}
\label{sec:implement}

\textbf{Network Architectures and Training.} For the anatomy generative prior, BoneFM, we parameterized the conditional velocity field with PointNet++~\cite{qi2017pointnet++} backbone modulated by FiLM layers~\cite{perez2018film}. BoneFM was trained on the 5,502 $P_{\text{AC}}$ training samples for 700 epochs, with $\approx$ 3.6 minutes per epoch, using the AdamW optimizer and an initial learning rate of $10^{-4}$. Classifier-free guidance (CFG) was used by randomly dropping the conditioning signal during training with probability $p_\text{drop} = 0.2$, with guidance weight $w = 2.0$ at evaluation. For the physics proxy, USimNet, we employed a PointNet++-based architecture. It was trained on the paired simulation dataset using the AdamW optimizer with a learning rate of $5 \times 10^{-5}$ to minimize the CD loss. 

\textbf{Guided Inference.} 
During the test-time inference, we discretized the continuous time $t \in [0, 1]$ into $K = 35$ integration steps. To enhance the conditional generation quality of BoneFM, we applied classifier-free guidance (CFG) with a scale of $w = 2.0$. For the time-dependent physics guidance schedule $\lambda(t)$, we set the activation threshold to $t_{\text{enable}} = 0.5$, the polynomial scaling factor to $\gamma = 3$, and the maximum guidance weight to $\lambda_{\text{max}} = 0.6$. For observation anchoring, we FPS-subsampled the input to a fixed subset of $N_\text{vis}$ anchor points and re-projected them onto their OT-FM path at every integration step, while the remaining points were generated freely. To ensure numerical stability during shape deformation, the analytical gradients from the physics proxy were strictly clipped at a threshold of $1.0$. All experiments were conducted on 40GB NVIDIA A100 GPUs. More details are provided in the Supplementary Material.

\subsection{Evaluation Metrics and Baselines}
\label{sec:metrics_baselines}

\textbf{Simulation Quality Metrics.} 
To assess simulation fidelity, we evaluated the set-level distributional similarity between the aligned in-vivo test set ($P_{\text{in-vivo, test}}$) and the simulated outputs produced by our method and by Gafencu \etal's method~\cite{gafencuShapeCompletionDark2024}, respectively, under an unpaired bootstrap protocol. Since no one-to-one ground-truth correspondence exists, we repeatedly drew random subsets of $X$ simulated point clouds (matching the in-vivo test set size) over 100 trials. All clouds were resampled to 8,192 points for fair comparison. We report the Minimum Matching Distance based on Chamfer Distance (MMD-CD $\downarrow$) and Coverage (COV $\uparrow$) to measure geometric proximity and mode coverage. Additionally, we computed the Density Jensen-Shannon Divergence (JSD $\downarrow$) on voxelized density distributions and the Sliced Wasserstein Distance (SWD $\downarrow$) to rigorously capture occupancy and global distribution mismatches.

\textbf{Completion Baselines.} 
We compared our method against several point cloud completion baselines, including {PoinTr}~\cite{yu2021pointr} (Transformer-based), {SVDFormer}~\cite{zhu2023svdformer} (Transformer-based), {PCDreamer}$^\dagger$~\cite{wei2025pcdreamer} (Adapted diffusion-based approach used as a multi-modality baseline), {Gafencu \etal's method}~\cite{gafencuShapeCompletionDark2024} (VAE-based), and one clinical statistical model, {SSM-Net$^*$}~\cite{cootes2024statistical}. We implemented SSM-Net$^*$, a parametric baseline inspired by SMPL~\cite{SMPL:2015}, which models vertebral shapes via PCA-derived blend shapes and linear blend skinning trained on the registered ground-truth meshes ($\mathcal{M}_{\text{GT, train}}$). For a fair evaluation, the resulting mesh outputs were uniformly converted into point clouds.

\textbf{Completion Metrics (Simulation Data).} 
 We evaluated all methods by taking $P_\text{phys-full}$ as input and comparing the completed output against the anatomical reference $P_\text{AC}$. Following recent works~\cite{yu2021pointr, wei2025pcdreamer}, we used $L1$ Chamfer Distance ($L1$ CD$\times 10^3\downarrow$), Earth Mover's Distance (EMD$\times 10^3\downarrow$), and F-score@1\%$\uparrow$ to quantitatively evaluate the completion results. All metrics were computed after normalizing both the predicted and reference point clouds to a unit sphere.

\textbf{Clinical Metrics (In-Vivo Data).} 
We report the laminae distance error (mm) measured by an experienced operator to evaluate the anatomical fidelity of UBone3D-completed vertebrae against both (1) the raw partial observations and (2) the VNN-reconstructed reference volumes. The completion results were compared to SVDFormer$\dagger$ (physically-trained) and SSM-Net$^*$.
\begin{figure}[!t]
    \centering
    \includegraphics[width=.95\linewidth]{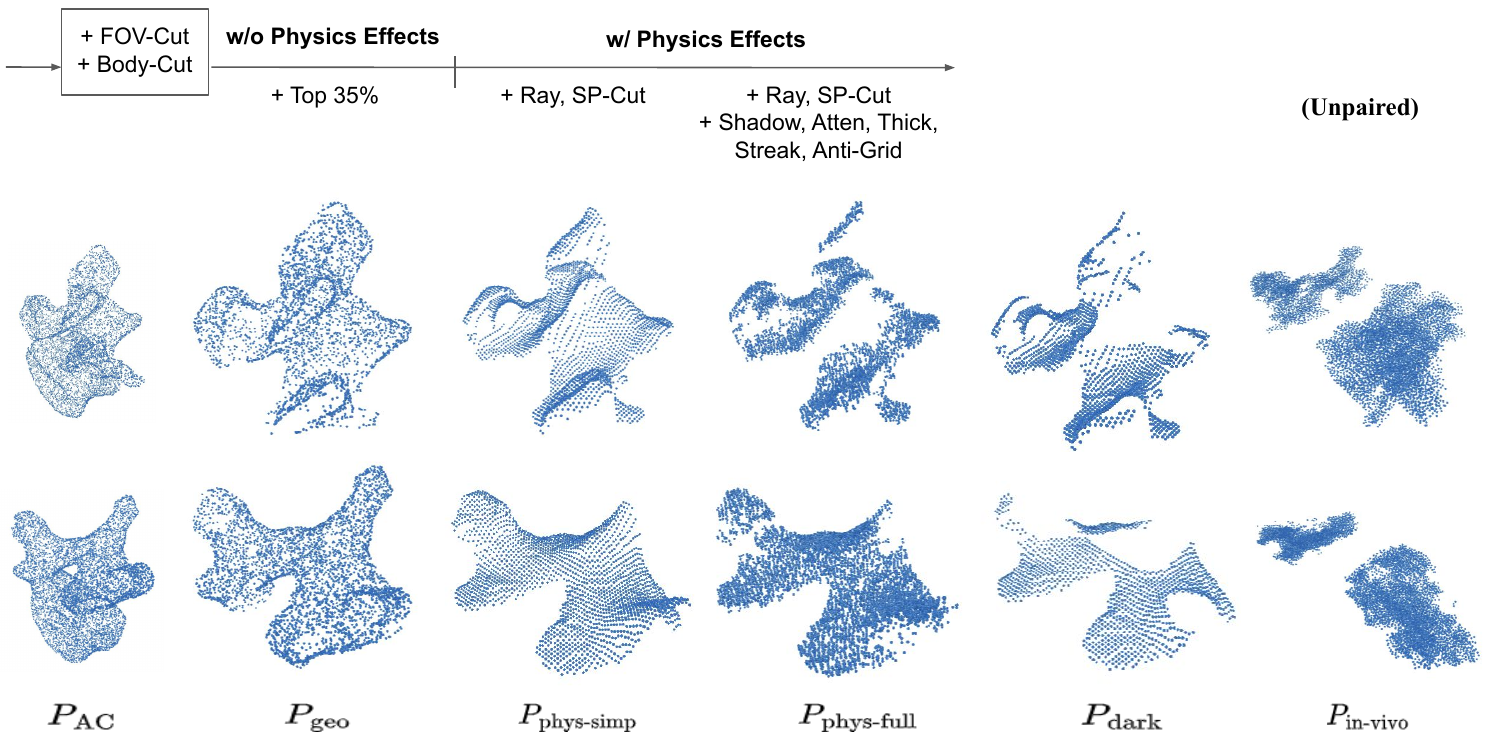}
    \caption{Comparison of simulation variants derived from the same $\mathcal{M}_{\text{GT}}$, all after applying a FOV-Cut and a Body-Cut. ${P}_\text{geo}$ is a purely geometric baseline obtained by shallow cropping with top-35\% points (Top 35\%), ${P}_\text{phys-simp}$ simulates minimal physics effects, i.e., first-return ray casting (Ray) and spinous-process removal (SP-Cut). ${P}_\text{phys-full}$ further adds acoustic shadowing (Shadow), depth windowing and exponential attenuation (Atten), thick-surface responses (Thick), axial streak artifacts (Streak), and anti-grid sampling perturbations (Anti-Grid: Poisson + jitter + subsampling + noise). The in-vivo examples (right) are shown for reference only and are not paired with the simulated point clouds.}
    \label{sim}
\end{figure}

% Implementation of PCDreamer$^\dagger$ and SSM-Net$^*$, and measurement of in-vivo laminae distance are detailed in the Supplementary Material.

\begin{table}[!t]
\centering
\caption{Set-level distributional similarity with respect to in-vivo data $P_\text{in-vivo}$.}\label{sim2real}
\resizebox{.6\textwidth}{!}{
\setlength{\tabcolsep}{6pt}
\begin{tabular}{@{}l|cccc@{}}
\toprule
\textbf{Data} & \textbf{MMD-CD$\downarrow$} & \textbf{COV$\uparrow$} & \textbf{JSD$\downarrow$} & \textbf{SWD$\downarrow$} \\
\midrule
$P_{\text{AC}}$        & 1298.480 & 0.124 & 0.664 & 16.246 \\
$P_{\text{geo}}$        & 961.705 & 0.140 & 0.649 & 14.083 \\
$P_{\text{dark}}$~\cite{gafencuShapeCompletionDark2024} & 970.563 & 0.179 & 0.655 & 14.293 \\
$P_{\text{phys-simp}}$  & 964.187 & 0.176 & 0.656 & 14.309 \\
$P_{\text{phys-full}}$  & \textbf{877.394} & \textbf{0.205} & \textbf{0.638} & \textbf{12.917} \\
\bottomrule
\end{tabular}
}
\end{table}

\section{Results and Discussions}

\subsection{Efficacy of the Physics Proxy}

Fig.~\ref{sim} provides a side-by-side visual comparison between in-vivo ultrasound point clouds ${P}_{\text{in-vivo}}$, a geometric baseline ${P}_{\text{geo}}$ and simulated point clouds, i.e., ${P}_{\text{phys-simp}}$ and ${P}_{\text{phys-full}}$. We also compared to point clouds $P_\text{dark}$ produced following Gafencu \etal's~\cite{gafencuShapeCompletionDark2024} method. Three distinct patterns are observed: (1) ${P}_{\text{geo}}$ and $P_\text{dark}$ mainly preserve a clean bone surface, (2) physics-driven structural missingness emerges in the simulated point clouds due to ultrasound view dependence and acoustic dropouts, and (3) ${P}_{\text{phys-full}}$ additionally exhibits realistic, structured artifacts. In particular, the full simulation captures key ultrasound characteristics which are commonly observed in real scans. The simulated point clouds show strong qualitative agreement with in-vivo data. Table~\ref{sim2real} shows the set-level distributional similarity between simulated and in-vivo point clouds ${P}_\text{in-vivo}$. The sampled CT point clouds ${P}_\text{AC}$, as the baseline, exhibit the largest distributional distance. Geometrically-cut ${P}_\text{geo}$ and simple physics ${P}_{\text{phys-simp}}$ show a closer match to in-vivo data. $P_\text{dark}$ does not exhibit a clear advantage over our settings and remains relatively far from the in-vivo distribution. Overall, our full physics simulation ${P}_{\text{phys-full}}$ shows the tightest statistical alignment with the in-vivo distribution. Taking both results into account, we conclude that ${P}_{\text{phys-full}}$ is a trustworthy proxy for real-world physics. 

\begin{figure*}[!t]
    \centering

    \begin{subfigure}{.9\linewidth}
        \centering
        \includegraphics[width=\linewidth]{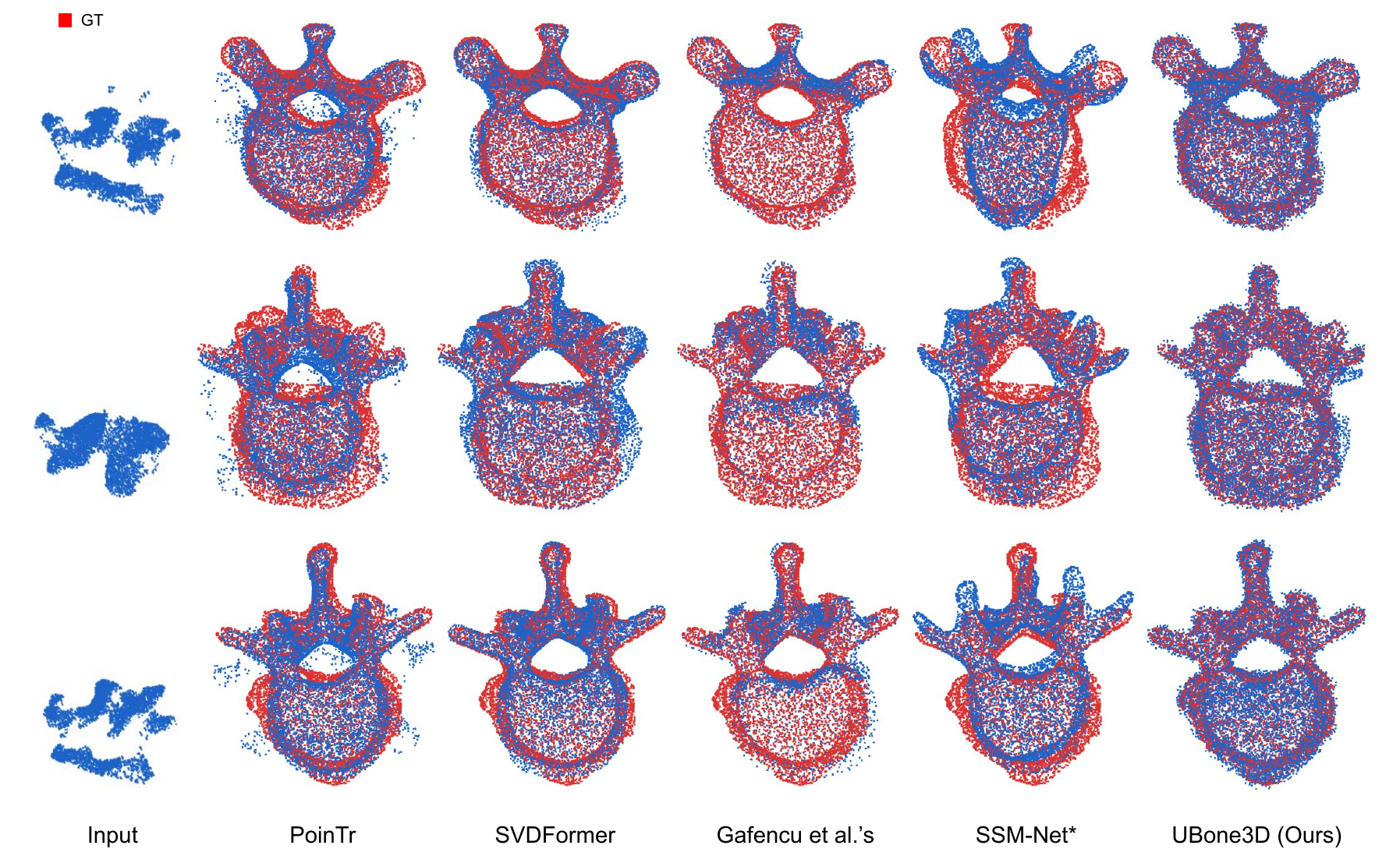}
        \caption{}
    \end{subfigure}
    \begin{subfigure}{.7\linewidth}
        \centering
        \includegraphics[width=\linewidth]{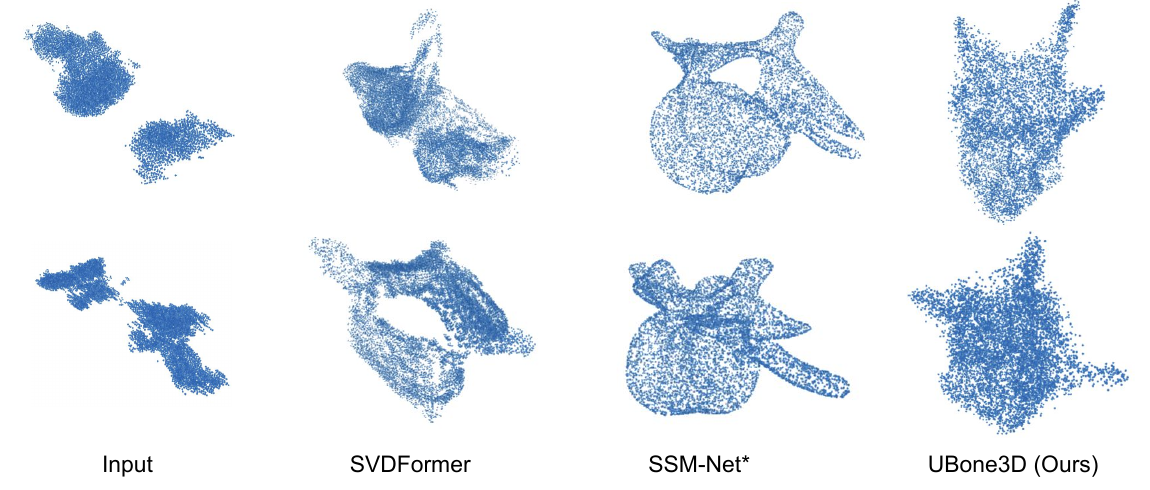}
        \caption{}
    \end{subfigure}
    \caption{Qualitative comparisons on simulation data (a) and in-vivo data (b). }   
    \label{res}
\end{figure*}
\subsection{Completion Results of Simulation Data} 
Fig.~\ref{res}~(a) presents qualitative visualization comparisons of UBone3D against baselines. Visually, PoinTr~\cite{yu2021pointr}, SVDFormer~\cite{zhu2023svdformer}, and SSM-Net$^*$ produced inferior results, exhibiting additional noise, lacking anatomical individuality, or generating the transverse process independently of the input. Although recovering generally good vertebral arch regions, Gafencu et al.'s method~\cite{gafencuShapeCompletionDark2024} produced a small number of points on the vertebral bodies. The adapted baseline, PCDreamer~\cite{wei2025pcdreamer}, unfortunately, failed to generate reasonable outputs. In contrast, our method preserved both global anatomical consistency and local structural sharpness. Table~\ref{tab2} reports the quantitative comparisons. Although Gafencu \etal's method and SVDFormer achieved the lowest CD under the two training settings, respectively, they exhibited substantially higher EMD values. Our framework shows consistently stable performance, suggesting that our method is better suited to handle ultrasound-specific artifacts than baselines.
% Please add the following required packages to your document preamble:
% \usepackage{booktabs}

% \usepackage[table,xcdraw]{xcolor}
% Beamer presentation requires \usepackage{colortbl} instead of \usepackage[table,xcdraw]{xcolor}
\begin{table}[!t]
\centering 
\caption{Quantitative comparisons on the simulation data with geometry-trained and simulation-trained variants ($L1$ CD$\times 10^3\downarrow$, EMD$\times 10^3$$\downarrow$, and F-score@1\%$\uparrow$).} \label{tab2}
\resizebox{.8\textwidth}{!}{
\setlength{\tabcolsep}{6pt}
\begin{tabular}{@{}l|ccc|ccc@{}} 
\toprule
\textbf{Method} & \textbf{CD$\downarrow$} & \textbf{EMD$\downarrow$} & \textbf{F1$\uparrow$} & \textbf{CD$\downarrow$} & \textbf{EMD$\downarrow$} & \textbf{F1$\uparrow$} \\ 
                & \multicolumn{3}{c|}{(trained with $P_\text{geo}$)} 
                & \multicolumn{3}{c}{(trained with $P_\text{phys-full}$)} \\
\midrule

PoinTr~\cite{yu2021pointr}           & 70.960 & {107.701} & 0.043 & 32.936 & 78.404  & 0.055 \\
SVDFormer~\cite{zhu2023svdformer}    & {42.106} & 207.075 & 0.033 & \textbf{25.463} & 127.965 & 0.046 \\
PCDreamer$\dagger$~\cite{wei2025pcdreamer} & 74.281 & 194.532 & 0.032 &   59.035    &    92.456     &   0.025    \\
Gafencu \etal~\cite{gafencuShapeCompletionDark2024}              & \textbf{41.648}  & 250.939  & 0.056 &   29.091    &     256.953    &    0.063   \\
SSM-Net$^*$              & 79.907  & 182.267   & 0.040 &   -    &     -    &    -   \\
\textbf{UBone3D} & 56.294  &  \textbf{105.292}  &  \textbf{0.065} &  28.161    &    \textbf{74.298}     &   \textbf{0.079}    \\ 
\bottomrule
\end{tabular}
} 
% \vspace{-2 mm}
\end{table}

\subsection{Completion Results of Zero-Shot In-Vivo Data}
The in-vivo qualitative and quantitative comparisons are provided in Fig.~\ref{res}~(b) and Table~\ref{lde}, respectively. Other baselines often failed to recover anatomical structure and produced distorted output. This might be due to their original design, which makes them struggle to generalize to versatile, artifact-dominated ultrasound inputs. Our method consistently produced anatomically plausible vertebrae with relatively correct bone structure observed from the visualization, and more accurate laminae distance as seen in Table~\ref{lde}. These results suggest that our model has potential for real-world patient data. Nevertheless, compared to the clear visual results on simulated data, the results on real-world (in-vivo) data may require further improvement.
\begin{table}[t]
\centering
\begin{minipage}[t]{0.55\textwidth}
\centering
\caption{Laminae Distance Errors (LDE) compared to raw input$^1$ and VNN reference$^2$.} \label{lde}
\resizebox{\textwidth}{!}{
\begin{tabular}[t]{@{}lccc@{}}
\toprule
& SVDFormer$\dagger$ &  SSM-Net$^*$ & \textbf{UBone3D} \\
\midrule
\textbf{LDE (mm)}$^1\downarrow$ & 0.838 & 1.253 & \textbf{0.819} \\
\textbf{LDE (mm)}$^2\downarrow$ & 3.937 & 3.382 & \textbf{1.293} \\
\bottomrule
\end{tabular}
}
\end{minipage}
\hfill
\begin{minipage}[t]{0.4\textwidth}
\centering
\caption{Ablation on the variants of physics proxy USimNet.} \label{tab:ablation_guidance}
% \vspace{-4mm}
\resizebox{\textwidth}{!}{
\begin{tabular}[t]{@{}lccc@{}}
\toprule
\textbf{Physics Proxy}& \textbf{CD$\downarrow$} & \textbf{EMD$\downarrow$} & \textbf{F1$\uparrow$} \\
\midrule
w/o USimNet      &         49.196         & 96.673 & 0.064 \\
Simple & 44.287       & 87.271 & 0.065 \\
\textbf{Full (Ours)} & \textbf{28.161}       & \textbf{74.298} & \textbf{0.079} \\
\bottomrule
\end{tabular}
}
\end{minipage}
\end{table}

\begin{figure}[t]
  \centering  
  \begin{minipage}{0.48\textwidth}
    \centering
\includegraphics[width=\textwidth]{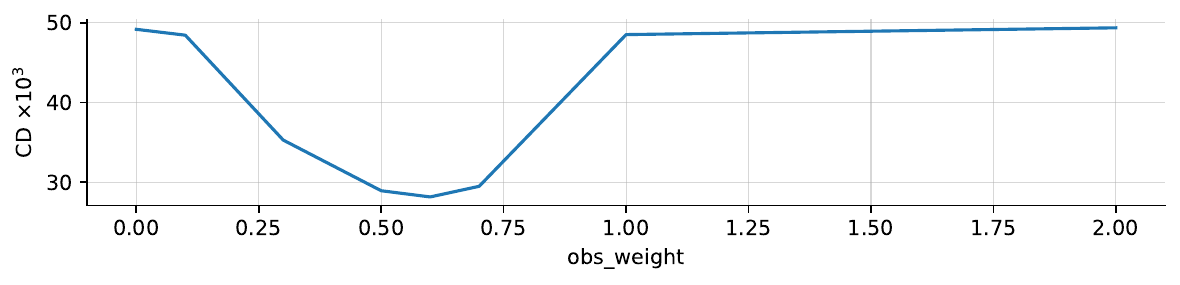}
    \captionof{figure}{Effect of the rectification strength.}
    \label{fig:obs_weight_a}
  \end{minipage}
  \hfill
  \begin{minipage}{0.48\textwidth}
    \centering
\includegraphics[width=\textwidth]{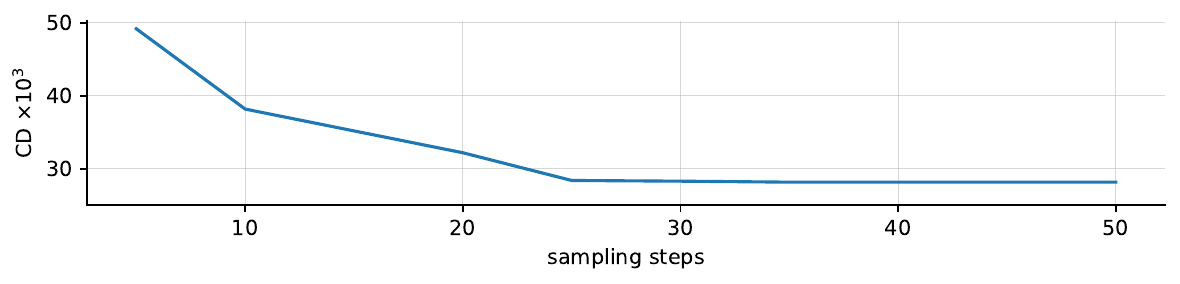}
    \captionof{figure}{Ablation on the sampling steps.}
    \label{fig:steps}
  \end{minipage}
\end{figure}
\begin{table}[!t]
\centering 
\caption{Ablation on the generative anatomy prior.} \label{tab:abdiffusion}
\resizebox{.6\textwidth}{!}{
\setlength{\tabcolsep}{6pt}
\begin{tabular}{@{}l|ccc|ccc@{}} 
\toprule
\textbf{Backbone}& \textbf{CD$\downarrow$} & \textbf{EMD$\downarrow$} & \textbf{F1$\uparrow$} & \textbf{CD$\downarrow$} & \textbf{EMD$\downarrow$} & \textbf{F1$\uparrow$} \\ 
                & \multicolumn{3}{c|}{(trained with $P_\text{geo}$)} 
                & \multicolumn{3}{c}{(trained with $P_\text{phys-full}$)} \\
\midrule

BoneDiff$\ddagger$& 59.274 & {105.920} & \textbf{0.068} & 42.597 & 75.987 & 0.071 \\
\textbf{BoneFM (Ours)}& \textbf{56.294}  &  \textbf{105.292}  & {0.065} &  \textbf{28.161}    &    \textbf{74.298}     &   \textbf{0.079}    \\ 
\bottomrule
\end{tabular}
} 
\end{table}

\subsection{Ablation Studies} \label{sec:ablation}
\textbf{Effectiveness of Physics Rectification.} An ablation study on the physics rectification variants is presented in Table~\ref{tab:ablation_guidance}. The BoneFM alone (w/o USimNet) yielded competitive completion results. Nevertheless, the introduction of physics rectification further improved completion fidelity. Simplified physics rectification (trained on $P_\text{phys-simp, train}$) stabilized the flow trajectory, while the full physics proxy (trained on $P_\text{phys-full, train}$) enforced the best physical plausibility. 

\textbf{Effectiveness of the Generative Anatomy Prior (\vs DDIM)}. We ablated the choice of generative model for the anatomy prior by comparing BoneFM to DDIM~\cite{song2020denoising} (BoneDiff$\ddagger$), each run at its own optimal number of sampling steps, with BoneDiff$\ddagger$ requiring 2$\times$ the steps of BoneFM to reach comparable convergence. As seen in Table~\ref{tab:abdiffusion}, BoneFM achieved better overall performance while requiring significantly fewer steps compared to BoneDiff$\ddagger$.

\textbf{Ablation on Rectification Strength.} Fig.~\ref{fig:obs_weight_a} illustrates the effect of varying the physics guidance weight $\lambda$ on reconstruction quality. The CD exhibited a U-shaped response, degrading under either insufficient or excessive guidance. The rectification strength of $\lambda = 0.6$ achieved the best balance between anatomy prior and physics rectification. $t_{\text{enable}}$ is fixed at 0.5.

\textbf{Ablation on Sampling Steps.} Fig.~\ref{fig:steps} shows the reconstruction quality with respect to the number of ODE integration steps, where $\lambda$ is fixed at 0.6. The performance increases markedly as the step count rises, with the most significant gains observed between 25 and 35 steps. To achieve a good trade-off between completion fidelity and inference efficiency, we set the number of steps to 35.

\section{Conclusion}
By leveraging an anatomical generative prior (BoneFM) and a differentiable physics proxy (USimNet), UBone3D completes anatomically consistent vertebral reconstructions from artifact-laden and partial ultrasound observations at inference time. Evaluations across simulation and in-vivo datasets confirm the effectiveness of our proposed framework. While further improvements are still needed, UBone3D provides a new perspective on clinical bone anatomy completion from ultrasound.

\section*{Acknowledgements}
The authors would like to thank all volunteers who participated in this study. This work was supported in part by the Women and Children’s Health Research Institute (WCHRI), Alberta Innovates, and the Natural Sciences and Engineering Research Council of Canada (NSERC). Ethics approval for this study was granted under protocol Pro00005707. 

% \clearpage  % TODO FINAL: This \clearpage needs to be removed from both review and camera-ready versions.
% ---- Bibliography ----
%
% BibTeX users should specify bibliography style 'splncs04'.
% References will then be sorted and formatted in the correct style.
%

\newpage
\bibliographystyle{splncs04}
\bibliography{main}
\end{document}